\documentclass[11pt]{article}

\usepackage{EMNLP2023}

\usepackage{times}
\usepackage{latexsym}
\usepackage{graphicx}
\usepackage{rotating}

\usepackage[T1]{fontenc}

\usepackage{microtype}

\usepackage{inconsolata}

\usepackage{multirow}

\title{LowResource at BLP-2023 Task 2: Leveraging BanglaBert for Low Resource Sentiment Analysis of Bangla Language}

  \author{Aunabil Chakma \\
  Bangladesh University of Engineering and Technology \\
  \texttt{0419052075@grad.cse.buet.ac.bd} \\\And
  Masum Hasan \\
  University of Rochester \\
  \texttt{m.hasan@rochester.edu} \\}

\begin{document}
\maketitle
\begin{abstract}

This paper describes the system of the LowResource Team for Task 2 of BLP-2023, which involves conducting sentiment analysis on a dataset composed of public posts and comments from diverse social media platforms. Our primary aim is to utilize BanglaBert, a BERT model pre-trained on a large Bangla corpus, using various strategies including fine-tuning, dropping random tokens, and using several external datasets. Our final model is an ensemble of the three best BanglaBert variations. Our system has achieved overall 3rd in the Test Set among 30 participating teams with a score of 0.718. Additionally, we discuss the promising systems that didn't perform well namely task-adaptive pertaining and paraphrasing using BanglaT5. Training codes and external datasets which are used for our system are publicly available at \href{https://github.com/Aunabil4602/bnlp-workshop-task2-2023}{https://github.com/Aunabil4602/bnlp-workshop-task2-2023}

\end{abstract}

\section{Introduction}

In the field of Natural Language Processing, Sentiment Analysis has earned significant attention as a research area dedicated to the analysis of textual content. A considerable body of research on Sentiment Analysis in Bangla has been conducted. Some of these works (e.g. \citet{islam-etal-2021-sentnob-dataset}, \citet{banglabook}) are based on introducing new datasets. In parallel, other works(e.g. \citet{vader}, \citet{sent-wvec-info}) are done on novel approaches. In spite of these numerous works, different opportunities still exist to improve the Analysis of Sentiments.

\begin{table}
    \centering
    \begin{tabular}{c c c}
        \hline
            \textbf{Rank} & \textbf{Team} & \textbf{Micro-f1} \\
        \hline
            1 & MoFa\_Aambela & 0.731 \\
            2 & yangst & 0.727 \\
            3 & LowResource(ours) & 0.718 \\
            4 & Hari\_vm & 0.717 \\
            5 & PreronaTarannum & 0.716 \\
        \hline
    \end{tabular}
    \caption{Showing top 5 of the final standings of the BLP-2023 Task 2. Our team stands 3rd among 30 participants.}
    \label{tab:final-rank-1}
\end{table}

In this paper, we describe our system for task 2 of the Bangla Language Processing Workshop @EMNLP-2023 \citep{blp2023-overview-task2}. We employ various systems based on BanglaBert and BanglaBert-Large \citep{bhattacharjee-etal-2022-banglabert}. Our experimental systems include fine-tuning, increasing the generalization based on dropping random tokens, using open-source external data during pre-training, and other methods described in section \ref{sec-sys-des}. Utilization of random token drop and external datasets has benefited our systems by improving micro-f1 scores around 0.006 to 0.01. Our best model, an ensembled model from three top models based on the Development Test-Set score, has scored a micro-f1 score of 0.7179, standing overall 3rd among 30 participants. Table \ref{tab:final-rank-1} shows the final standings of the task.

Additionally, we describe alternate potential methods that have not scored well in the result section \ref{results}. To illustrate, we explore Task Adaptive Pre-Training \citep{tapt-dapt}, in fact, has been used by this year's winner of SemEval Task 12 \citep{semeval-task12} on sentiment analysis of African Language, and generating paraphrases using BanglaT5 \citep{banglat5}. Moreover, we notice a significant drop in our score in the final test set of our best model. We describe this as our limitations in the section \ref{sec-limit}.

\section{Related Works}

Many of the related works are primarily focused on novel datasets covering diverse domains. \citet{emonoba} have developed a dataset comprised of various public comments from social media platforms. \citet{rahman-dey} have created their datasets based on Cricket and Restaurant reviews. Most recently, \citep{banglabook} have published a dataset entirely comprised of book reviews from online bookshops. 

Existing approaches to Sentiment Analysis on Bangla Language primarily rely on machine learning and deep learning techniques. For example, \citet{svm-based} have used Support Vector Machine(SVM) for Sentiment Analysis on public opinions on Cricket. Recurrent Neural Network based models are also highly used. (e.g. \citet{lstm}). \citet{triptovai} have explored a variety of approaches including LSTM, SVM, and Naive Bayes. Moreover, convolutional neural network based models are also used for Sentiment Analysis on Bangla(e.g. \citet{cnn-based}).

In recent years, Large Language Models(LLM), trained on huge corpus, have become popular for their capability to understand the language and can easily fine-tuned for any task like Sentiment Analysis. LLMs based on the Bangla language(e.g. BanglaBert \citep{bhattacharjee-etal-2022-banglabert}, shahajBert \citep{diskin2021distributed}, BanglaT5 \citep{banglat5}) are also available, which opens opportunities to work on various tasks for Bangla.

\section{Task Description}
 This is a multi-class classification task where the objective is to detect the sentiment of the given text into 3 different classes: Positive, Negative, and Neutral. The score will be calculated using the micro-f1. The task consists of two phases: a development phase followed by a test phase. The final standing is based on the score of the test set provided during the test phase.

\subsection{Dataset Description} The dataset is comprised of MUBASE \citep{hasan2023zero} and SentiNob \citep{islam-etal-2021-sentnob-dataset} datasets. The SentiNob dataset consists of various public comments collected from social media platforms. It covers 13 different domains, for example, politics, education, agriculture, etc. On the other hand, the MUBASE dataset consists of posts collected from Twitter and Facebook. The sample sizes of different sets given for training, validation, and testing are shown in Table \ref{tab:dataset-size}.

\begin{table}
    \centering
    \begin{tabular}{c c}
        \hline
            \textbf{Set Name} & \textbf{Sample size}\\
        \hline
            Training & 32566\\
            Development & 3934\\
            Development Test & 3426\\
            Test & 6707\\
        \hline
    \end{tabular}
    \caption{Sample sizes of various sets provided in the Task 2.}
    \label{tab:dataset-size}
\end{table}

\section{System Description}
\label{sec-sys-des}
Here, we discuss several systems that we have experimented with for the task including the pre-processing of the dataset.

\subsection{Fine-tuning Pre-trained LLMs} 
Fine-tuning Pre-trained Models can achieve high scores with fewer training steps. Top competitors of different shared tasks (e.g. \citet{ner-damo}, \citet{sa-nlnde}) use these pre-trained models. For this task, we use several variations of BanglaBert for fine-tuning. Namely,  we use BanglaBert and BanglaBert-Large. Besides, we also use XLM-Roberta-Large \citep{model-xlmr}, a multi-lingual model. We don't explore much on multi-lingual models, since we have found that monolingual models are more used than multi-lingual models on monolingual-specific tasks \citep{semeval-task12} due to high scores.

\subsection{Task Adaptive Pre-training of LLMs} 
\citet{tapt-dapt} suggest that Domain Adapting Pre-Training(DAPT) And Task Adaptive Pre-Training(TAPT) improve the scores of the corresponding task. Here, we do TAPT on BanglaBert and BanglaBert-Large using the Electra pre-training method \citep{electra}, which was originally used to pre-train these models. We don't perform DAPT since the models already cover the domains.

\subsection{2-Stage Fine-Tuning of LLMs} 
In the first stage, we fine-tune BanglaBert using the external data only. Here, we don't include any given data from the task. In the next stage, we do regular fine-tuning on the train set. We use the term "2FT" as a short form of this approach. The list of the external datasets and sample sizes are shown in table \ref{tab:extdata-list}.

\subsection{Data augmentation} 
We experiment with 2 data augmentation techniques to improve the generalization. First, instead of dropping random words \citep{data-aug}, we drop random tokens(RTD) since dropping words might change the meaning. We apply RTD on the fly during the training. Second, we employ paraphrasing as data augmentations using BanglaT5 \citep{banglat5}.

\subsection{Preprocessing of Data} 
We remove the duplicates found in the training set and development set. We replace any url and username with URL and USER tag respectively similar to \citet{bertweet}. While using BanglaBert we normalize the sentence by their specific normalizer\footnote{https://github.com/csebuetnlp/normalizer} as required by their model. All of the sentences are tokenized by the individual tokenizer required by each model. We set the max length of tokenization to 128 for each text. 

We use several external data. However, most of the labels don't match the labels of this task. For the initial fine-tuning of the LLMs, we first map different labels to the three labels for this task. The label mapping is shown in table \ref{tab:label-conv}. For TAPT, we didn't need any of these labels since we do masked language modeling. Finally, we also remove the duplicates found in the external datasets.

\section{Experimental Setup} 
We have used Models and Trainer from Huggingface\footnote{https://huggingface.co/}(PyTorch version). We employ mixed precision training \citep{amp} that enables faster training and consumes low GPU memory. Moreover, we built a code such that the results are reproducible. All of the experiments are done using a single V100 GPU in Google Colaboratory\footnote{https://colab.research.google.com/}. We do hyper-parameters search on learning rate, batch size, dropout ratios, and total epochs. We start the search with the parameter settings as suggested \citet{tapt-dapt}. Our best training parameters of fine-tuning and TAPT are shown in the table \ref{tab:hp-ft} and \ref{tab:hp-tapt} respectively. Note that, we don't use samples from the development set, development-test set, and test set for fine-tuning and pre-training.

\section{Results}
\label{results}

To begin, we discuss the systems that have scored well on the Development-Test's score. The top individual model is BanglaBert-Large with a random token drop that has scored 0.733, and even without any enhancement, it can score 0.723. The next best single model is BanglaBert with random token drop(RTD) and 2-stage fine-tuning that has scored 0.729. Table \ref{tab:devtest-result} shows the scores of our selected models in the Development-Test Set. Here, we see that both usages of external datasets and RTD have benefited the BanglaBert and BanglaBert-Large. We have built an ensemble of 3 best individual models(model ID 3, 5, and 6) that has scored 0.734, where we decide the class based on majority voting, and in case of a tie, we use the class predicted by the best model. We chose only the 3 best models for the ensemble because the other model's score was low and taking an odd number of models helps to decide the output class in case of a tie.

We have submitted the ensembled model as our best model in the test phase and has scored 0.718. Moreover, We have submitted the 3 individual best models. Our scores on the Test Set are shown in table \ref{tab:test-result}. Here, we have found some inconsistency: BanglaBert-Large with random token drop, which we have considered the best model based on the Development-Test set, performed worst among the other 2 models, and BanglaBert with random token drop and pre-fine-tuned with external data, our 2nd best model, has performed the best. More importantly, every variant of BanglaBert-Large has scored low on the Test set. We discuss some analysis more in section \ref{sec-limit}. Finally, table \ref{tab:confusion-matrix} shows the confusion matrix of our ensembled model on Test set. We see that our model performed worst on detecting the Neutral class, i.e. only 412 out of 1277 samples have been correct having an accuracy of 32\%, where the accuracy of Positive and Neutral classes are 78\% and 83\% respectively.

There are some systems that didn't achieve favorable performance from the beginning of our experiments. Firstly, TAPT didn't improve our results but rather declined the score by 0.039 with respect to simple fine-tuning as shown in table \ref{tab:tapt-para-result}. What we can infer is that TAPT is supposed to help adapt the BanglaBert to the task domain, but it overfitted on the training samples, where the original model is already in a good optima that covered the task domain better. 

\begin{figure}[ht]
    \begin{center}
        \includegraphics[scale=0.5]{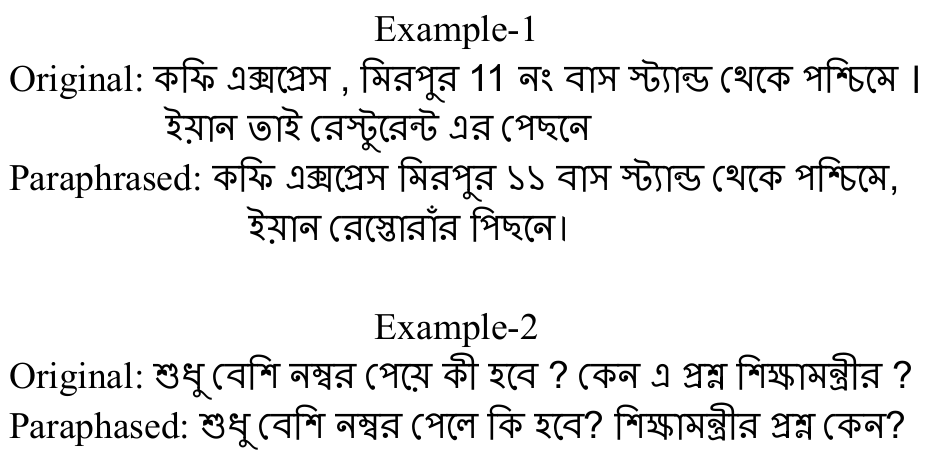}
        \caption{Showing 2 paraphrasing examples using BanglaT5.}
        \label{pic:para-banglat5}
    \end{center}
\end{figure}

Paraphrasing to create additional data using BanglaT5 also didn't work well. Its score is shown in table \ref{tab:tapt-para-result}. The most perceptible reason is that paraphrased sentences, although good, were not diverse enough from the original sentences. Examples of generated paraphrases are shown in figure \ref{pic:para-banglat5}.

Other than BanglaBert, we try the XLM-Roberta-Large, a multi-lingual model, which is used by several task winners (e.g. \cite{ner-damo}). However, it has scored low on the Development-Test set even with all enhancements. Its score is also shown in Table \ref{tab:devtest-result}.

\begin{table}[h]
    \centering
    \begin{tabular}{c c c}
        \hline
            \textbf{ID} & \textbf{System} & \textbf{Micro-F1}\\
        \hline
            1 & BBert & 0.718\\
            2 & BBert+RTD & 0.722\\
            3 & BBert+RTD+2FT & 0.729\\ 
            4 & BBertL & 0.723\\
            5 & BBertL+RTD & 0.733\\
            6 & BBertL+RTD+2FT & 0.725\\
            7 & XLM-Roberta-Large+RTD & 0.713 \\ 
            8 & Ensemble(3+5+6) & 0.734\\
        \hline
    \end{tabular}
    \caption{Scores of our best models of various systems in the Development-Test Set. Here, BBert means BanglaBert, RTD means random token drop and 2FT means 2-stage fine-tuning.}
    \label{tab:devtest-result}
\end{table}

\begin{table}[h]
    \centering
    \begin{tabular}{c c}
        \hline
            \textbf{System} & \textbf{Micro-F1}\\
        \hline
            BBertL+RTD & 0.711\\
            BBert+RTD+2FT & 0.719\\
            BBertL+RTD+2FT & 0.713\\ 
            Ensemble(above 3 models) & 0.718\\
        \hline
    \end{tabular}
    \caption{Scores of 3 of our best models(based on Development-Test Set) and ensembled model in Test Set. Here, BBert means BanglaBert, RTD means random token drop and 2FT means 2-stage fine-tuning.}
    \label{tab:test-result}
\end{table}

\begin{table}
    \centering
    \begin{tabular}{c c}
        \hline
            \textbf{System} & \textbf{Micro-F1}\\
        \hline
            Fine-Tuning &  0.727\\
            TAPT &  0.688\\
            Paraphrasing & 0.674\\ 
        \hline
    \end{tabular}
    \caption{Performance of TAPT and Paraphrasing on BanglaBert-Large in comparison with fine-tuning on Development Set.}
    \label{tab:tapt-para-result}
\end{table}

\begin{table}
\begin{tabular}{l|l|c|c|c|}
\multicolumn{2}{c}{}&\multicolumn{3}{c}{Predicted}\\
\cline{3-5}
\multicolumn{2}{c|}{}&Neg&Neut&Pos\\
\cline{2-5}
\multirow{3}{*}{
\begin{turn}{90}
   True
\end{turn}
}& Neg & 2770 & 244 & 324\\
\cline{2-5}
& Neut & 598 & 412 & 267\\
\cline{2-5}
& Pos & 331 & 128 & 1633\\
\cline{2-5}
\end{tabular}
\caption{Confusion Matrix of the Ensembled model on Test Set.}
    \label{tab:confusion-matrix}
\end{table}

\section{Limitations and future work}
\label{sec-limit}
As mentioned earlier, we find inconsistency in the score of our best model (BanglaBert-Large) between the Development-Test Set and the Test Set. \citet{electra} have stated that variance in performance is observed with different seeds when the size of the dataset is small. We assume this might be the cause, although we didn't rely on other seeds since the distribution of Development-Test Set and Test Set should be similar as they come from the same datasets. To be more certain, we ran an experiment using different seeds for both BanglaBert and BanglaBert-Large on the Test set. As anticipated, models show varying performance when initialized with different seeds. Table \ref{tab:seed-incro} shows the results of this experiment. Moreover, we have found that the average score of the BanglaBert is better than the BanglaBert-Large. In fact, this result is consistent with the result found by the authors of BanglaBert that BanglaBert-Large performs lower than BanglaBert on Sentiment Analysis on SentiNob dataset\footnote{https://huggingface.co/csebuetnlp/banglabert\_large}. BangalThus, before considering a model, the average score from different seeds needs to be evaluated when the training data is small.

\begin{table}
    \centering
    \begin{tabular}{c c c}
        \hline
            \textbf{Seed} & \textbf{BBert} & \textbf{BBertL}\\
        \hline
            1234 & 0.7156 & 0.7115\\
            42 & 0.7179 & 0.7110\\
            747 & 0.7197 & 0.7210\\ 
            52467 & 0.7192 & 0.7122\\
            2779 & 0.7135 & 0.7161\\
            362 & 0.7185 & 0.7134\\
            8194 & 0.7182 & 0.7127\\ 
        \hline
            avg. & 0.7177 & 0.7140\\
        \hline
    \end{tabular}
    \caption{Scores from using different seeds for BanglaBert(BBert), BanglaBert-Large(BBertL) on Test Set.}
    \label{tab:seed-incro}
\end{table}

TAPT is a popular method for pre-training, but it has been ineffective for our task. However, we have inferred this based on a few experiments. Thus, we suggest that more research needs to be done on the effectiveness of TAPT, as well as DAPT, on BanglaBert.  

Our research has been mostly based on fine-tuning. As future work, we would like to explore using common data augmentation techniques \citep{data-aug} for the given data. Besides, there are several multilingual Pre-trained Models that include the Bangla Language are need to be explored along with sophisticated methods and may even achieve better results.

\section{Conclusion}

In this paper, we stated our systems based on BanglaBert and BanglaBert-Large for that Sentiment Analysis task. We used simple techniques like, 2-stage fine-tuning, using external datasets, and dropping random tokens. Our system scored 3rd overall in the task. We also discussed some potential systems that didn't demonstrate satisfactory performance. More importantly, we have discussed the score inconsistency of our best model between Development-Test Set and Test Set as our limitation. Finally, we discussed directing some future research like applying TAPT and DAPT on BanglaBert and trying more data augmentations or sophisticated methods.

\bibliography{anthology,custom}
\bibliographystyle{acl_natbib}

\appendix

\section{Appendix}
\label{sec:appendix}

Here, we show a figure and additional tables related to our descriptions.

\begin{table}[h]
    \centering
    \begin{tabular}{c c c}
        \hline
            \textbf{Parameter} & \textbf{BBert(+L)} & \textbf{XLMR}\\
        \hline
            Learning Rate(LR) & 2e-5 & -\\
            LR Scheduler & Linear & -\\
            Warmup Ratio & 0.0 & -\\
            Train Batch Size & 16 & 32 \\
            Train Epochs & 3 & 5\\
            Weight Decay & 0.01 & -\\
            Token Drop Ratio & 0.2 & -\\
            Classifier Dropout & 0.1 & - \\ 
            Max Length & 128 & - \\
        \hline
    \end{tabular}
    \caption{Best hyper-parameter settings for fine-tuning. BBert(+L) means both the BanglaBert and BanglaBertLarge models, XLMR means the XLM-Roberta-Large model, and "-" means equal to the left column values.}
    \label{tab:hp-ft}
\end{table}

\begin{table}[h]
    \centering
    \begin{tabular}{c c}
        \hline
            \textbf{Parameter} & \textbf{BBertL}\\
        \hline
            $\lambda$ & 50\\
            MLM probability & 0.25 \\
            Learning Rate(LR) & 1e-4\\
            LR Scheduler & Linear\\
            Warmup Ratio & 0.06\\
            Train Batch Size & 64\\
            Train Epochs & 100\\
            Weight Decay & 0.01\\
            Token Drop Ratio & 0.2\\
            Max Length & 128 \\
        \hline
    \end{tabular}
    \caption{Best hyper-parameter settings for Electra Pre-Training of BanglaBert and BanglaBert-Large during Task Adaptive Pre-training(TAPT). $\lambda$ is the weight of the loss for Discriminator}
    \label{tab:hp-tapt}
\end{table}

\begin{table}[h]
    \centering
    \begin{tabular}{c c c}
        \hline
            \textbf{Dataset} & \textbf{Samples} & \textbf{Total class}\\
        \hline
            \citet{emonoba} & 22739 & 6\\
            \citet{sazzed2021abusive} & 1000 & 2\\
            \citet{bemoc} & 7000 & 6\\
            \citet{banglabook} & 158065 & 3\\
            \citet{sazzed2020cross} & 11807 & 2\\
            \citet{rahman-dey} & 2979 & 3\\
            \citet{omar-srw} & 4994 & 6\\
            \citet{ban-absa} & 9014 & 3\\
            \citet{khondoker} & 14852 & 3\\
        \hline
    \end{tabular}
    \caption{The list of External Datasets used for our training.}
    \label{tab:extdata-list}
\end{table}

\begin{table}[h]
    \centering
    \begin{tabular}{c c}
        \hline
            \textbf{Original} & \textbf{Converted}\\
        \hline
            Love & Positive\\
            Joy & Positive\\
            Anger & Negative\\
            Sad/Sadness & Negative\\
            Fear & Negative\\
            Disgust & Negative\\
            Surprise & Neutral\\
            Abusive & Negative\\
            Non-Abusive & Positive\\
        \hline
    \end{tabular}
    \caption{Label conversions of external datasets for aligning to our task.}
    \label{tab:label-conv}
\end{table}

\end{document}